# Fuzzy Logic of Speed and Steering Control System for Three Dimensional Line Following of an Autonomous Vehicle

Dr. Shailja shukla
Department of Electrical Engineering
Jabalpur engineering College Jabalpur
Jabalpur (M.P) India
E-mail shailja.shukla@indiatimes.com

Mr. Mukesh Tiwari
Department of Electrical Engineering
Jabalpur Engineering College Jabalpur
Jabalpur (M.P) India
E-mail mukesh_tiwari836@yahoo.co.in

*Abstract*- The major problem of robotics research today is that there is a huge barrier to entry into Robotics research due to system software complexity and need for a researcher to learn more about details, dependencies and intricacies of the complete system. This is because a robot system needs several different modules to communicate and execute in parallel. Today there are not much controlled comparisons of algorithms and solutions for a given task, which is the standard scientific method of other sciences. There is also very little sharing between groups and projects, requiring code to be written from scratch over and over again. This paper is to describe exploratory research on the design of a modular autonomous mobile robot controller. The controller incorporates a fuzzy logic [8] [9] approach for steering and speed control [37], a FL approach for ultrasound sensing and an overall expert system for guidance. The advantages of a modular system are related to portability and transportability, i.e. any vehicle can become autonomous with minimal modifications. A mobile robot test bed has been constructed in university of Cincinnati using a golf cart base. This cart has full speed control with guidance provided by a vision system and obstacle avoidance using ultrasonic sensors. The speed and steering fuzzy logic controller is supervised through a multi-axis motion controller. The obstacle avoidance system is based on a microcontroller interfaced with ultrasonic transducers. This micro-controller independently handles all timing and distance calculations and sends distance information back to the fuzzy logic controller via the serial line. This design yields a portable independent system in which high speed computer communication is not necessary. Vision guidance has been accomplished with the use of CCD cameras judging the current position of the robot.[34] [35][36] It will be generating a good image for reducing an uncertain wrong command from ground coordinate to tackle the parameter uncertainties of the system, and to obtain good WMR dynamic response.[1] Here we Apply 3D line following mythology. It transforms from 3D to 2D and also maps the image coordinates and vice versa, leading to the improved accuracy of the WMR position. The fuzzy logic Controller may give a good command signal; moreover we can find a highly accurate plant model to design the controller taking into account

The unknown factors like friction and dynamic environment .This design, in its modularity, creates a portable autonomous fuzzy logic controller applicable to any mobile vehicle with only minor adaptations.

## I. INTRODUCTION

Controller design for any system needs some knowledge about the system. Usually this involves a mathematical description of the relation among inputs to the process, its state variables, and its Output. This description is called the model of the system. The model can be represented as a set of transfer functions for linear time invariant systems or other relationships for non-linear or time variant systems.

Modeling of complex systems may be very difficult task. In a complex system such as a multiple input and multiple outputs system inaccurate models can lead to unstable systems, or unsuitable system performance. [9] Fuzzy Logic Control (FLC) is an effective alternative approach for systems which are difficult to model. The FLC uses the qualitative aspects of the human decision process to construct the control algorithm. This can lead to a robust controller design. The modeling of a mobile robot is a very complex task and a direct application of FLC can be found in this area. An excellent introduction to the mathematical analysis of mobile robots can be found in [1]

Even though the visualization and recognition of image information for the guidance of mobile robot have been studied for many years, The design of a mobile vehicle system is a challenging task in the sense determining what information to measure and how to use this information to design an intelligent controller in a manner that will satisfy the performance specifications of the system.

Overall fuzzy logic approaches for modeling control systems for vehicles have been studied in the past. A fuzzy logic controller [8] that guarantees stability of a control system for a computer simulated model car and advanced fuzzy logic application for automobiles application has been discussed in Altrock et. al. [9].

## II. OBJECTIVES

The main aspect of intelligent control addressed in this paper is the design of a controller for a mobile robot using fuzzy logic. The design specifications selected here fully satisfies the building of a robot simulation which could follow a line, avoid obstacles, and adapt to variations in terrain. The adaptive capabilities of mobile robots depend on the fundamental analytical and architectural designs of the sensor systems used.

The mobile robot provides an excellent test platform for investigations into generic vision guided robot control since it is similar to an automobile and is a multi-input, multi-output system. An algorithm has been developed to establish a mathematical and geometrical relationship between the physical three dimensional (3-D) ground coordinates of the line to follow and its corresponding two dimensional (2-D) digitized image coordinates.

This relationship is incorporated into the vision tracking system to determine the perpendicular distance and angle of the line with respect to the centroid of the robot. The information from the vision tracking system is used as input to a closed loop fuzzy logic controller to control the steering and the speed of the robot.

## III. RESEARCH OBJECTIVE

The main goal of this research is to model a modular Fuzzy Logic Control for an automated guided vehicle and test the performance of the vehicle Simulation in A MATLAB Simulation the research is focused on the design of the Fuzzy Controller for vision and sonar navigation of the automated guided vehicle.

The design of the controller has been executed in three stages. In the first stage the universe of discourse is identified and fuzzy sets are defined. The rule base (Fuzzy Control Rules) for the control is then defined through a human decision making process. The membership functions and their intervals are defined. Aggregation and de fuzzification methods are selected. In the second stage the Fuzzy Controller is implemented on the autonomous guided vehicle. In the third and final stage performance of the controller is tested through a series of simulations and real time running of the vehicle.

## IV. METHODOLOGY

The purpose of the vision guidance is to guide the robot to follow the line using a digital charge couple device (CCD) camera. To do this, the camera needs to be calibrated. Camera calibration 34 is a process to determine the relationship between a given 3-D coordinate system (world coordinates) and the 2-D image plane a camera perceives (image coordinates). More specifically, it is to determine the camera and lens model parameters that govern the mathematical or geometrical transformation from world coordinates to image coordinates based on the known 3-D control field and its image. The CCD camera digitizes the line from 3-D coordinate system to 2-D image system. Since the process is autonomous, the relationship between the 2-D system and the 3-D system has to be accurately determined so that the robot can follow the line. The objective of this section is to show how a model was developed to calibrate the vision system so that, given any 2-D image coordinate point, the system can mathematically compute the corresponding ground coordinate point. The X and Y (the Z is constant) coordinates of two ground points are then computed from which the angle and the perpendicular distance of the line with respect to the centroid of the robot are determined. The vision system was modeled by the following equations

$xPI = A_{11} X_g + A_{12} Y_g + A_{13} Z_g + A_{14}$ ..............(1)

$yPI = A_{21} X_g + A_{22} Y_g + A_{23} Z_g + A_{24}$ ..............(2)

Where $A_{nm}$ are coefficients, xPI and yPI are x and y image coordinates, and $X_g$, $Y_g$, and $Z_g$ are the ground coordinates. In transforming the ground coordinate points to the image coordinate points the following

### IV.II, 2-D TRANSFORMATION OPERATION

- Translation
- Scaling
- Rotation
- Shear

### IV.II.I, 2D TRANSLATION

- Current position

$$P = \begin{bmatrix} x \\ y \end{bmatrix}$$

- New position (after translation)

$$P = \begin{bmatrix} x' \\ y' \end{bmatrix}$$

- Translation operation

$$T(\delta x, \delta y) = \begin{bmatrix} \delta x \\ \delta y \end{bmatrix}$$

$p + T(\delta x, \delta y) = p'$ ............. (3)

$x' = x + \delta x$ ................. (4)

$y' = y + \delta y$ ............... (5)

### IV.II.II 2, SCALING

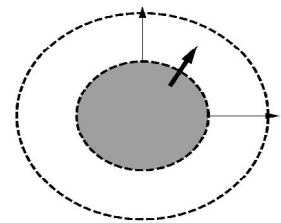

**Fig. 1** Scaling

- Current position

$$P = \begin{bmatrix} x \\ y \end{bmatrix}$$

- New position

$$P' = \begin{bmatrix} s_x x \\ s_y y \end{bmatrix}$$

- Scaling operation

$$S(s_x, s_y) = \begin{bmatrix} s_x & 0 \\ 0 & s_y \end{bmatrix}$$

$S(s_x, s_y) p = p'$ ....... (6)

- Matrix multiplication

### IV.II.III 2D-ROTATION

- Current position

$$P = \begin{bmatrix} x \\ y \end{bmatrix}$$

- New position

$$P = \begin{bmatrix} x' \\ y' \end{bmatrix}$$

- Rotation operation

$$R(\theta) = \begin{bmatrix} \cos(\theta) & -\sin(\theta) \\ \sin(\theta) & \cos(\theta) \end{bmatrix}$$

$$R(\theta)\mathbf{p} = \mathbf{p}' \quad \ldots\ldots\ldots (7)$$

$$P' = \begin{bmatrix} x\cos(\theta) - y\sin(\theta) \\ x\sin(\theta) + y\cos(\theta) \end{bmatrix}$$

- Positive angles are "counter-clockwise"!

- Derivation of rotation

$$x = r\cos(\theta_1) \quad \ldots\ldots (8)$$
$$y = r\sin(\theta_1) \quad \ldots\ldots (9)$$

- Rotate $\theta_2$

$$x' = r\cos(\theta_1 + \theta_2) \quad \ldots\ldots (10)$$
$$y' = r\sin(\theta_1 + \theta_2) \quad \ldots\ldots (11)$$

- Observation (important results from trigonometry)!

$$x' = r\cos(\theta_1)\cos(\theta_2) - r\sin(\theta_1)\sin(\theta_2) \quad \ldots\ldots (12)$$
$$y' = r\cos(\theta_1)\sin(\theta_2) + r\sin(\theta_1)\cos(\theta_2) \quad \ldots (13)$$

2D Rotation

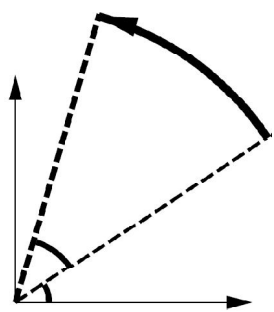

**Fig. 2. 2D** Rotation

**IV.II.IV, 2D SHEAR**

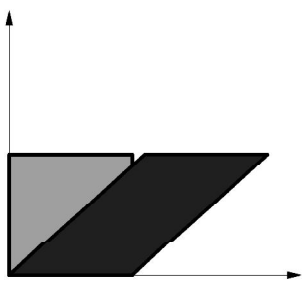

**Fig. 3.** 2D Shear

- Current position

$$P = \begin{bmatrix} x \\ y \end{bmatrix}$$

$$\mathbf{p}' = \begin{bmatrix} x' \\ y' \end{bmatrix} = \begin{bmatrix} x + ay \\ y \end{bmatrix}$$

- Shear position

$$Sh_x(a) = \begin{bmatrix} 1 & a \\ 0 & 1 \end{bmatrix}$$

$$\mathbf{p}' = Sh_x(a)\mathbf{p} \quad \ldots\ldots\ldots (14)$$

- Geometric meaning
- Shear operation along y axis

$$Sh_y(b) = \begin{bmatrix} 1 & 0 \\ b & 1 \end{bmatrix}$$

$$\mathbf{p}' = Sh_y(b)\mathbf{p} = \begin{bmatrix} x \\ bx + y \end{bmatrix}$$

- Geometric meaning!
- Consider more complicated cases!
- Various examples are shown in the class

Transformation operations occur on the points: scaling, translation, rotation, perspective, and projective. Solving for the transformation parameters to obtain the image and ground coordinate relationship is a difficult task. Fortunately, in the model equations given above, the transformation parameters are embedded into the coefficients. To compute the coefficients, a calibration device was built to obtain 12 data points. With the 12 points, a matrix equation was yielded as Shown below

$$XPI = CA_{1K} \quad \ldots\ldots\ldots (15)$$
$$YPI = CA_{2K} \quad \ldots\ldots\ldots (16)$$

Where

$$C = \begin{bmatrix} xg_1 & yg_1 & zg_1 & 1 \\ xg_2 & yg_2 & zg_2 & 1 \\ \cdot & \cdot & \cdot & \cdot \\ \cdot & \cdot & \cdot & \cdot \\ \cdot & \cdot & \cdot & \cdot \\ xg_{12} & yg_{12} & zg_{12} & 1 \end{bmatrix}; XPI = \begin{bmatrix} xPI_1 \\ xPI_2 \\ \cdot \\ \cdot \\ \cdot \\ xPI_{12} \end{bmatrix}; YPI = \begin{bmatrix} yPI_1 \\ yPI_2 \\ \cdot \\ \cdot \\ \cdot \\ yPI_{12} \end{bmatrix}; A_{1k} = \begin{bmatrix} A_{11} \\ A_{12} \\ A_{13} \\ A_{14} \end{bmatrix}; A_{2k} = \begin{bmatrix} A_{21} \\ A_{22} \\ A_{23} \\ A_{24} \end{bmatrix}$$

Eqns. (15) & (16) consist of 12 linearly independent equations and four unknowns; the least-square regression method is applied to yield a minimum mean square error solution for the coefficients. Below are the equations for the solution:

$$A_{1K} = (C^T C)^{-1} C^T \text{XPI} \ldots \ldots \ldots \ldots (17)$$

$$A_{2K} = (C^T C)^{-1} C^T \text{YPI} \ldots \ldots \ldots \ldots (18)$$

Given an image coordinate xPI and yPI, and z ground coordinate (the z coordinate of the points with respect to the centroid of the robot is maintained constant since the robot is run on a flat surface in this model) the corresponding $X_g$ and $Y_g$ ground coordinates are computed as indicated by the following matrix equations.

$$\begin{pmatrix} x_g \\ y_g \end{pmatrix} = Q^{-1} B \ldots \ldots \ldots \ldots (19)$$

Where

$$Q = \begin{bmatrix} A_{11} & A_{12} \\ A_{21} & A_{22} \end{bmatrix}; B = \begin{bmatrix} xPI - A_{14} - (A_{13} z_g) \\ yPI - A_{24} - (A_{23} z_g) \end{bmatrix}$$

Note that equation (1) and (19) can be modified to accommodate the computation of $Z_g$ when an elevation of the ground surface is considered. The image processing of the physical points is done by the ISCAN tracking device, which returns the centroid of the brightest or darkest region in a computer controlled windows and returns its X and Y coordinates. Two points on the line are windowed and their corresponding coordinates are computed as described above. From the computed x and y ground coordinates of the points, the angle of the line with respect to the centroid of the robot is computed from simple trigonometric relationship. In the next section, we shall show how the angle of the line just computed is used with other parameters to model the steering control of the robot with a fuzzy logic controller.

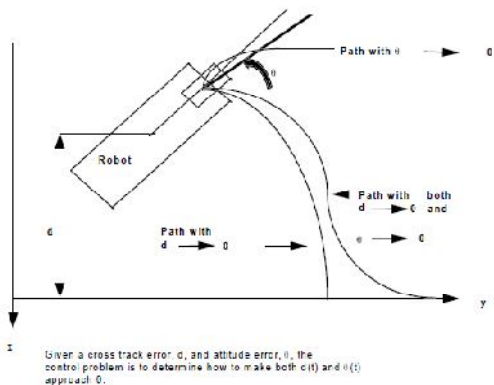

**Fig. 4** Line following

## V. OBSTACLE AVOIDANCE

The obstacle avoidance system consists of six ultrasonic transducers. An ultrasonic ranging system from Polaroid is used for the purpose of calibrating the ultrasonic transducers. An Intel 80C196 microprocessor and a circuit board with a liquid crystal display is used for processing the distance calculations. The distance value is returned through a RS232 port to the control computer. The system requires an isolated power supply: 10-30 VDC, 0.5 amps. The two major components of an ultrasonic ranging system are the transducer and the drive electronics. In the operations of the system, a pulse of electronic sound is transmitted toward the target and the resulting echo is detected. The elapsed time between the start of the transit pulse and the reception of the echo pulse is measured. Knowing the speed of sound in air, the system can convert the elapsed time into a distance measurement. The drive electronics has two major categories - digital and analog. The digital electronics generate the ultrasonic frequency. A drive frequency of 16 pluses per second at 52 kHz is used in this application.

## VI. DESCRIPTION OF THE AUTONOMOUS VEHICLE

The system that is to be controlled is an electrically propelled mobile vehicle was created and assembled during the spring quarter of 1998 in the Advanced Robotics Lab at the University of Cincinnati. This vehicle was built as part of the Autonomous Guided Vehicle contest sponsored by the Army Tank Command. A 3D rendering of above said mobile vehicle is shown below. The vehicle is constructed of an aluminum frame designed to hold the controller, obstacle avoidance, vision sensing, vehicle power system, and drive components. Two independently driven brushless DC motors are used for both vehicle propulsion, as well as for vehicle steering This independent drive system not only gives vehicle the capability to move in a straight line, and perform turning actions, but this system also allows the vehicle to have a zero turning radius feature. This feature allows the vehicle to turn directly about the center of the drive without requiring forward motion, thereby giving the vehicle the ability to navigate through more complicated course requirements. A 3-axis Gallil motion control board is used as the interface between the controller CPU and the drive components, including the brushless servo drive motors and the encoders. A Galill board, the brushless motor, and an optical encoder provide a closed loop system that allows the controller code to specify accurately motor dynamics parameters, including position, velocity and acceleration. The Gallil controller contains a digital PID type controller. This controller is tuned with the Servo Design Kit Software package, that selects the Proportional, Integral, and Derivative gains (Kp, Ki, Kd) to optimize the system response.

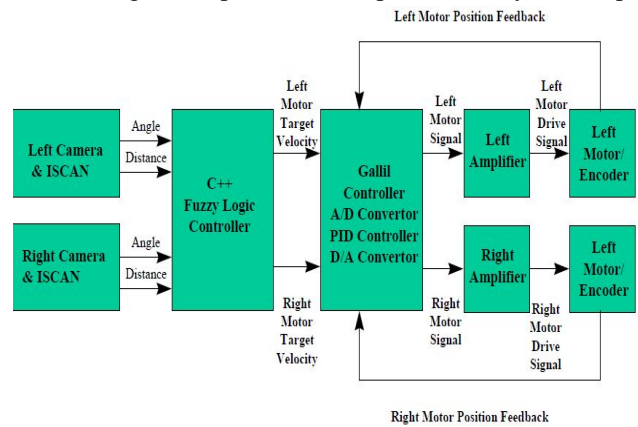

**Fig. 5** Description of the Autonomous Vehicle

Once tuned, the controller code is able to simply select motor position, velocity and acceleration, and control the trajectory of the mobile Vehicle. It is important to distinguish these PID components from the control logic used for vehicle: this controller drives the motor to a specified velocity, whereas the control logic selects the value of that specified velocity based on vision system inputs. There are two control system types that may be used to control an autonomous guided vehicle. The firm is to provide the vehicle with complete information about the environment, and the required path.

The vehicle then uses navigation sensors, and the programmed information about the environment to navigate through the course this method requires extensive programming to completely define the path, and is unable to navigate in an unknown territory. The second method is to gather information from the environment using external sensors, and process the information to control the speed and steering parameters of the vehicle. Due to the unknown environment for the usage of the AGV, this second method is utilized. For path generation, the input to the controller is attained through two cameras mounted on the front of the vehicle. These cameras sense the visual image of the path-line. This information is processed through the controller

### VII. SIMULATION MODEL OF AN AUTONOMOUS VEHICLE

In the simulation model of the autonomous vehicle there are six inputs and three output all these six inputs are combined by using multiplexer and provide single input to fuzzy logic controller. All the inputs evaluated for each rule that create the fuzzy rule editor and give the single output which is decoded by de-multiplexer. Two outputs for left and right motor speed are closed loop with feedback follow the characteristic of the PID and unity gain. This measured error signal between the desired and actual state of the linear closed loop control system, is driven to a zero value and the desired state is achieved

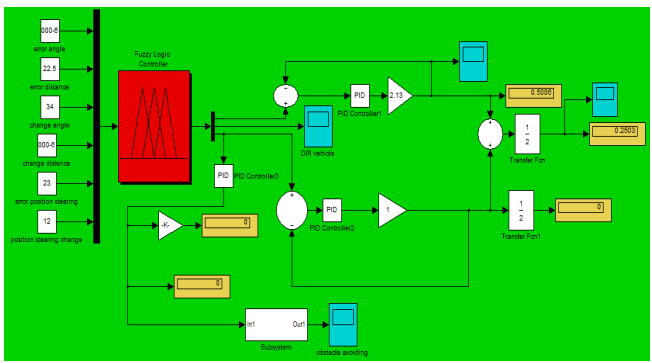

**Fig. 6** Simulation model of An Autonomous Vehicle

### VIII. ISCAN SYSTEM

The ISCAN system returns two variables: the distance of the vehicle from the border line and the relative angle between the path of the vehicle and the border line. The controller utilizes these two inputs to select motor speeds that will drive the vehicle to follow the specified path without crossing the boundary lines. The vehicle can be driven in a straight line by specifying equivalent angular velocities of the motors, or driven in a turn by specifying difference between angular velocities of the motors. A schematic of the components of the control system is through a fuzzy logic algorithm, the code translates the input from the vision system into target output velocities for each of the right and left vehicle drive motors.

These outputs are processed through the Gallil motion control system that translates the target output into controller output signals. These signals are passed through the amplifiers to increase the signals to a level that will drive the motors at the proper velocity. The encoders, mounted on the ends of the motors, supply an angular position feedback signal back to the Galill control board to minimize the steady state error between the target velocity, and the actual motor velocity. The end result is the desired vehicle motion at the target velocity. This system provides motor output signals for each input sample. The system is therefore able to adjust to a changing environment, and changing path conditions[1] .The previous control logic for the AGV utilized a classical linear control system with feedback, and a PID (Proportional, Integrator, Derivative) controller.

This system has been selected due to the simplicity of its components. This system proved to be excellent for challenges focusing mainly on the complexity of the multiple input / multiple output, and the difficulty of tuning the system with respect to changing environments. For this reason, a fuzzy logic controller, which is simpler to understand, and to modify, has been created. To control the vehicle, a control algorithm is developed that can easily be coded for the Gallil control hardware interface. The Gallil board translates speed commands into output signals which, after proper tuning, result in a desired motor angular position, velocity and acceleration. The membership functions were tuned to improve the performance of the vehicle.

### IX. FUZZY LOGIC CONTROLLER

Fuzzy logic controller [9] [13] uses the fuzzy set and fuzzy logic theory previously introduced in its implementation. A detailed reference on how to design a fuzzy controller can be found in 29, 30, and 31. Fuzzy Inference System Fuzzy inference is the actual process of mapping from a given input to an output using fuzzy logic 27. Fuzzy logic starts with the concept of a fuzzy set. A fuzzy set is a set without a crisp, clearly defined boundary. It can contain elements with only a partial degree of membership. The MATLAB Fuzzy Logic Toolbox was used to build the initial experimental Input fuzzy sets the first level of the fuzzy system has two inputs, error and error. These inputs are resolved into a number of different fuzzy linguistic sets.

### X. RULE BASE

The way one develops control rules [16] depends on whether or not the process can be controlled by a human operator. If the operator's knowledge and experience can be explained in words, then linguistic rules can be written immediately. If the operator's skill can keep the process under control, but this skill cannot be easily expressed in words, then control rules may be formulated based on the observation of operator's actions in terms of the input - output operating data. However, when the process is exceedingly complex, it may not be controllable by a human expert. In this case, a fuzzy model of the process is built and the control rules are derived theoretically.

It should be noted however, that this approach is quite complicated and has not yet been fully developed. Therefore the FLC is ideal for complex ill - defined systems that can be controlled by a skilled human operator without the knowledge of their underlying dynamics. In such cases an FL controller is quite easy to design and implementation is less time consuming than for a conventional controller
.

### XI. RESULTS

The testing of the dual level fuzzy system controller explained in that has been done in two steps. First, a theoretical simulation is run using the MATLAB fuzzy tool box.

The input for this simulation has generated using theoretical test cases. The inputs to the MATLAB batch file which is use to run the simulation (M-file) were error and error. The outputs were steering and the speeds of the motor two cases were considered. Results of the simulation are shown in Obstacle avoidance with case staid.

1. Straight line path.

The objective of this case is to test if the controller handles a simple case as straight line this simulation is presented in Figure.1 The output indicating a successful line following in this path has all condition are zero settling time , angle, peak time.

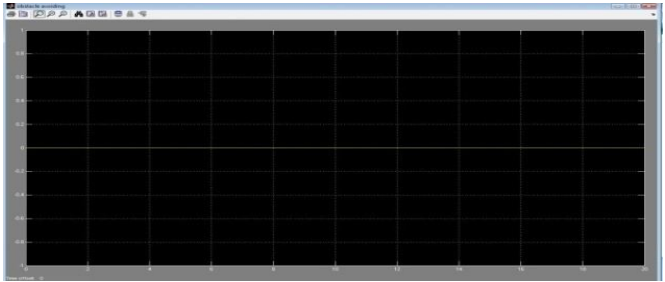

**Fig. 7** Straight line path

2. Curved Path.

In Case 2, the robot Is made to follow a curved line. In this case the input data was free of any noise in form of obstacles and loss of vision. The curved path direction of vehicle graph as stating time 0sec. Settling angle 35.settling time 20sec.The result of this simulation as seen in Figure 2 suggests that the fuzzy inference system was able to direct the vehicle along curved lines.

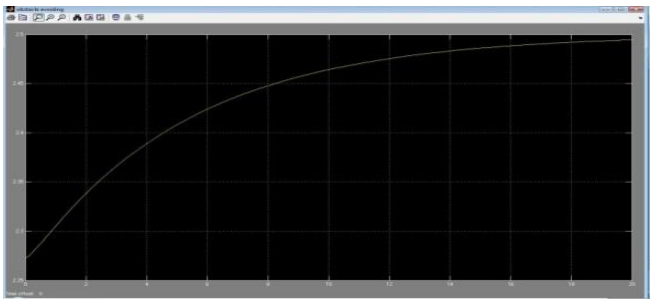

**Fig. 8** Curved Path

3. Angular path.

In Case 3, the robot followed a curved line. Noise in form of loss of input vision data and obstacle is used during data collection stage. Angular path conditions has starting time 2.2 sec settling angle 35 settling time 19sec.The simulation presents a successful line following. The inference engine worked successfully for this case

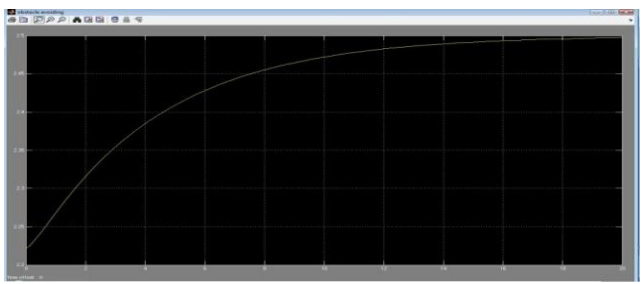

**Fig. 9** Angular fluctuations path

4. Extreme environment path.

In Case 4, an extreme environment with excessive noise in form of adverse vision input data and obstacles in sensitive positions (on curves as pointed out in the simulation) have been introduced. To make this case more difficult steep curves were used. The simulation presents a failure at the point when the robot path crosses the line it is supposed to follow. This extreme environment path graph has starting time of obstacle simulation 0 sec.,settling angle is 60 degree and settling tine 14sec.

This presents a limitation of the fuzzy inference engine. This limitation arises due to the limitation on the number of rules that can be implemented in this present controller. The FLC fails due to the excess adverse parameters introduced in this case. A solution to this problem is to use a neural network to identify such extreme conditions and use dedicated fuzzy inference engines for each case.

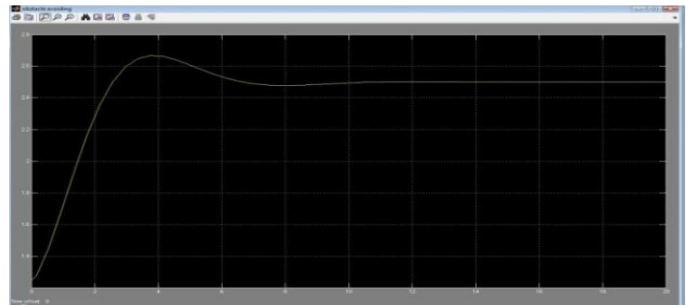

**Fig. 10** Environment path

TABLE.1

| Direction of Vehicle | Obstacle avoidance | | |
|---|---|---|---|
| | Starting Time sec. | Settling angle θ | Settling time sec. |
| Straight Line Path | 0 | 0 | 0 |
| Curved Path | 0 | 35 | 20 |
| Angular Path | 2.2 | 35 | 19 |
| Environment Path | 0 | 60 | 14 |

## XII. MOTOR SPEED

Components, including the brushless servo drive motors and the encoders. A Galill board, the brushless motor, and an optical encoder provide a closed loop system that allows the controller code to specify accurately motor dynamics parameters, including position, velocity and acceleration. The Gallil controller contains a digital PID type controller.

This controller is tuned with the Servo Design Kit Software package, that selects the Proportional, Integral, and Derivative gains (Kp, Ki, Kd) to optimize the system response. In right motor torque 0.6N/ms in t= 2 sec., peak time torque is 0.7N/ms in t=2.4sec. and settling time torque is 0.5N/ms in t= 16 sec. for left motor speed as well as right motor speed.

Once tuned, the controller code is able to simply select motor position, velocity and acceleration, and control the trajectory of the AGV,. It is important to distinguish these PID components from the control logic used for vehicle: this controller drives the motor to specified velocity, whereas the control logic selects the value of that specified velocity based on vision system inputs has the code for the M-file. Note that the results appear reasonable for both angle and speed. The results of these simulations encouraged further tests using a real life scenario. As a result the model has been also implemented on the mobile robot, which is scheduled to take part in nationwide obstacle avoidance and part following competition

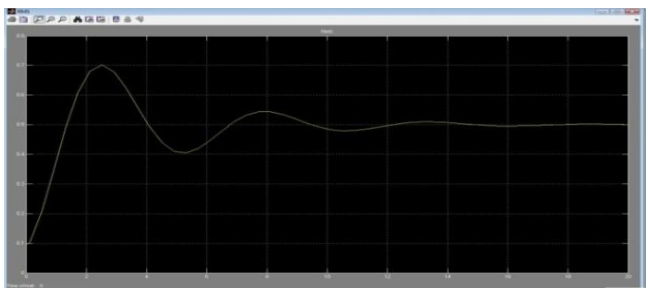

**Fig. 11** Right motor speed

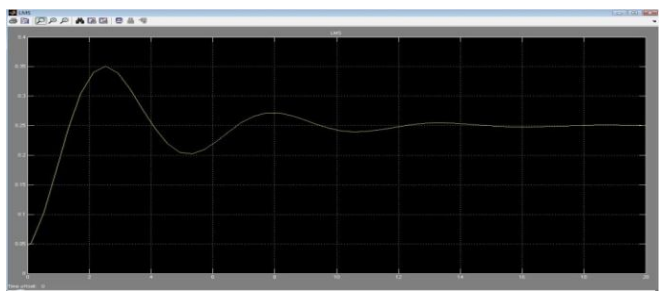

**Fig. 12** Left motor speed

**TABLE.2**

| Motor | $T_r/t_r$ | $T_p/t_p$ | $T_s/t_s$ |
|---|---|---|---|
| Right motor | 0.6/2 | 0.7/2.4 | 0.5/16 |
| Left motor | 0.6/2 | 0.7/2.4 | 0.5/16 |

## XIII. CONCLUSION

The design and implementation of a modular fuzzy logic based controller for an autonomous mobile robot for line following along with position control with respect to an obstacle course has been presented. The control algorithm for this application is based on vision navigation. The development of the [8] [9] FLC controller was accomplished after the detailed study of an autonomous guided vehicle and its environment. A rule base was generated using expert system knowledge. Fuzzy membership functions and fuzzy sets were developed. The FLC model was first tested on the MATLAB fuzzy logic toolkit with some special cases. A number of tests were run to analyze the stability and response of the system under fuzzy control in a real life scenario. Tuning of the system in form of adjusting the membership functions and the rules has been accomplished to improve the stability of the FLC. The fuzzy logic control is a very flexible and robust soft computing tool for control. The number of variants involved in the current application present a challenge for any type of control system. A fuzzy logic control was selected, as a soft computing solution for this problem keeping in minds its robustness and flexibility. The performance of the robot was studied with simulations for five different cases selected for the study. The FLC shows good stability and response for three of the cases. The problem at hand seems to be a complex problem for just one inference engine to handle. This limitation arises due to the limit on the number of rules and membership functions that can be used in a single inference engine. A better system performance can be obtained if a FL approach is used. The environment in which the robot runs should be divided into a number of specific classes according to input data. The control system model will contain an identified and classified input data, which will finally fire the right inference engine for the input data class. From the results obtained in the MATLAB simulation and the preliminary testing of the model on the robot, it can be concluded that the model presented, can be reliably and successfully implemented permanently on the robot. Fuzzy logic has been proven to be an excellent solution to control problems where the number of rules for a system are finite and which can be easily established.[16] In this application an infinite number of rules can also be established. The fuzzy control in a way acts as a learning system control, as it has the ability to learn from situations where it fails. This learning is possible by increasing the number of rules in the system. In this way the system can keep on learning until it becomes a perfect system.

AUTHORS PROFILE

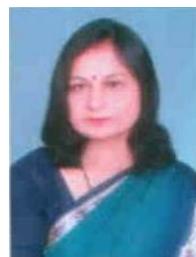

*Dr. Shailja Shukla* received B.E. degree in Electrical Engg. from Jabalpur Engg. College, Jabalpur in 1984 and the Ph.D. degree in Control System from Rajiv Gandhi Technical University, Bhopal in 2002. She is currently Professor in Electrical Engg. and the Chairperson of the Department of Computer Science and Engg. at Jabalpur Engg. College, Jabalpur. Her research interest on Large Scale Control Systems, Soft Computing and include Machine Learning, Face Recognition and Digital Signal Processing. She has been the Organizing Secretary of International Conference on Soft Computing and Intelligent Systems. She has published more than 40 Research papers in International/National Journals and Conferences. She is Editorial member of many International Journals

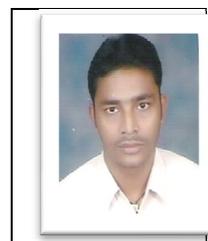

*Mr. mukesh tiwari* was born in Madhya-Pradesh at katni on 27th November 1983. He received B.E degree in (Electronic & Communication) from Rewa Institute of Technology Rewa in 2007 he is the student of master of Engineering in (Control system) Department of Electrical Engineering. Jabalpur Engineering College Jabalpur (M.P) INDIA. His research interest on fuzzy logic, communication, and control system He has published one International journal & Three National Conference